\documentclass[letterpaper, 10 pt, conference]{ieeeconf}  % Comment this line out if you need a4paper

\IEEEoverridecommandlockouts                              % This command is only needed if 
\title{\LARGE \bf
3D Multi-Object Tracking with Semi-Supervised GRU-Kalman Filter
}

\author{Xiaoxiang Wang$^{1}$, Jiaxin Liu$^{2}$, Miaojie Feng$^{3}$, Zhaoxing Zhang$^{4}$ and Xin Yang$^{*}$ % <-this % 
\thanks{*Corresponding author}
\thanks{$^{1}$Xiaoxiang Wang, $^{2}$Jiaxin Liu, $^{3}$Miaojie Feng, $^{4}$Zhaoxing Zhang and $^{*}$Xin Yang are with School of Electronic Information and Communications, Huazhong University of Science and Technology, Wuhan, 430074, China.
        {\tt\small \{xiaoxiang0012,ljx123,fmj,zzx,xinyang2014\}@hus
        t.edu.cn}}%
}

\usepackage{amsmath}

\usepackage{booktabs}
\usepackage{threeparttable} 

\usepackage{array}

\usepackage{algorithmic}
\usepackage{algorithm}
\usepackage{bm}

\usepackage{hyperref}
\usepackage{makecell}

\usepackage{cite}

\usepackage{graphicx}

\begin{document}

\maketitle
\thispagestyle{empty}
\pagestyle{empty}

%%%%%%%%%%%%%%%%%%%%%%%%%%%%%%%%%%%%%%%%%%%%%%%%%%%%%%%%%%%%%%%%%%%%%%%%%%%%%%%%
\begin{abstract}

3D Multi-Object Tracking (MOT), a fundamental component of environmental perception, is essential for intelligent systems like autonomous driving and robotic sensing. Although Tracking-by-Detection frameworks have demonstrated excellent performance in recent years, their application in real-world scenarios faces significant challenges. Object movement in complex environments is often highly nonlinear, while existing methods typically rely on linear approximations of motion. Furthermore, system noise is frequently modeled as a Gaussian distribution, which fails to capture the true complexity of the noise dynamics. These oversimplified modeling assumptions can lead to significant reductions in tracking precision. To address this, we propose a GRU-based MOT method, which introduces a learnable Kalman filter into the motion module. This approach is able to learn object motion characteristics through data-driven learning, thereby avoiding the need for manual model design and model error. At the same time, to avoid abnormal supervision caused by the wrong association between annotations and trajectories, we design a semi-supervised learning strategy to accelerate the convergence speed and improve the robustness of the model. Evaluation experiment on the nuScenes and Argoverse2 datasets demonstrates that our system exhibits superior performance and significant potential compared to traditional TBD methods.The code is available at \href{https://github.com/xiang-1208/GRUTrack}{https://github.com/xiang-1208/GRUTrack}.

\end{abstract}

%%%%%%%%%%%%%%%%%%%%%%%%%%%%%%%%%%%%%%%%%%%%%%%%%%%%%%%%%%%%%%%%%%%%%%%%%%%%%%%%
\section{INTRODUCTION}

 Multi-Object Tracking (MOT)\cite{weng20203d} is a crucial research topic within the field of computer vision and serves as a foundational technology in numerous intelligent applications, such as autonomous driving, traffic flow analysis, security surveillance, robotics, and action recognition.

At present, with the increasing performance of Multi-Object Detection (MOD)\cite{yin2021center,redmon2016you,carion2020end,chen2023voxelnext}, Multi-Object Tracking (MOT) methods based on the ``Tracking-by-Detection'' (TBD) \cite{weng20203d, pang2022simpletrack, li2023poly} have demonstrated superior accuracy and robustness. These TBD methods follow the motion process of the tracked object, contrasting with ``Joint Detection and Tracking'' (JDT) approaches \cite{huang2021joint,yin2021center,zaech2022learnable} which do not perform as well. In general, TBD methods update the state of tracked objects incrementally by constructing motion model and employing recursive Bayesian filter estimator. However, due to the varied motion characteristics of different objects within the scene, a single state space (SS) and estimator parameter cannot well match the different motion characteristics between various categories, which reduces the consistency between the motion state update and the actual, leading to false matching and inaccurate state update. Some approaches\cite{li2023poly,kim2021eagermot,wang2023camo} take note of this and design motion parameters or association strategies for each class to be more relevant to the different characteristics of different classes.

However, these methods fail to fundamentally solve the problem of multi-category differences. On the one hand, with the continuous addition or refinement of categories, it is not only tedious to design motion models for each category, but also too dependent on the designer's experience. On the other hand, although it is possible to capture different types of motion characteristics by continuously refining the categories, most methods still use model-based state filters, such as Kalman Filter (KF)\cite{Kalman} and Extended Kalman Filter (EKF)\cite{Gruber,larson1967application}. The effectiveness of these models depends on the accuracy of the state model and the validity of the motion hypothesis. In the actual MOT, the latent state of the system is nonlinear and complex, and it is even difficult to be accurately described as a tractable state model. In this case, model-based state estimators typically simplify the motion dynamics by linearizing the process and assuming the system noise follows a Gaussian distribution. This assumption does not match the actual situation, and these modeling inaccuracies often bring the loss of tracking precision to the system.

To this end, we propose a partially learnable MOT method by introducing a Gated Recurrent Unit (GRU)-based Kalman filter into the motion module of TBD. This method can replace the traditional manual model design in a data-driven manner, thus eliminating the need to design a unique SS and estimator for each class. Specifically, we use multiple GRUs to simulate the loops in Recursive Bayesian Filtering. The model automatically learns the noise distribution, state transition matrix and observation matrix. It avoids the mismatch of noise modeling and the loss of precision caused by linearizing the state transition and observation function. This is doable in theory because neural network-based state estimation has been shown to capture the motion characteristics of complex processes\cite{chung2014empirical,graves2012long}, which is also applicable to state transitions in MOT.

\begin{figure*}[tbp]
    \centering
	\includegraphics[width=1\linewidth]{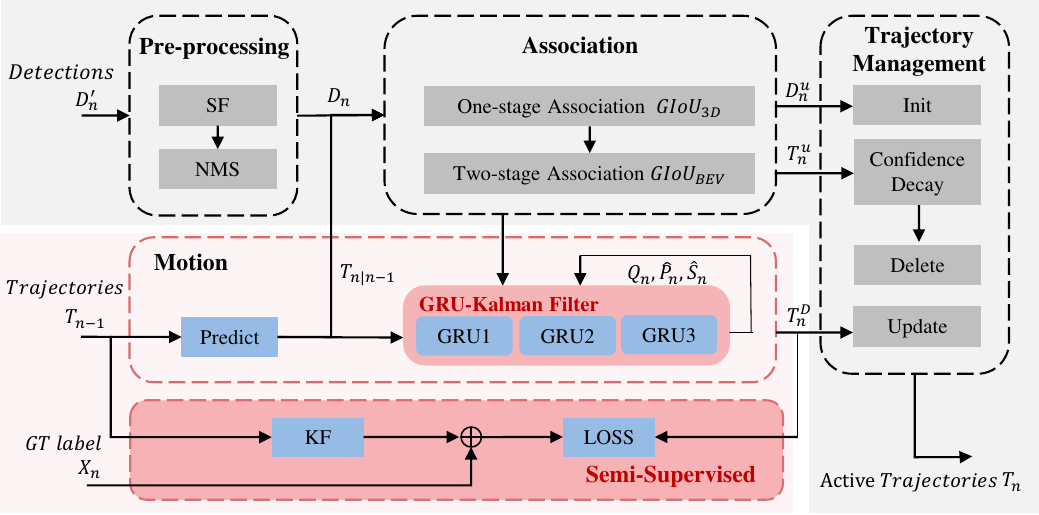}
    \caption{\textbf{The pipeline of our proposed method at frame n.} $T^D_n$ is the trajectories updated by associating upper observations $D_n$ and using the motion module. Our design focuses on two parts, one is \textbf{GRU-Kalman Filter}: it uses three GRUs to simulate the Kalman filtering process. The second is \textbf{Semi-Supervised learning}, which uses dataset annotations and pseudo-labels generated by a parallel Kalman filter for joint training.}
    \label{fig:Overview}%文中引用该图片代号
\end{figure*}

However, it is not feasible to directly use the learnable Kalman Filter in MOT. On the one hand, due to the partial annotation of the dataset, the amount of trainable data is small, which is easy to overfitting. On the other hand, since the annotations and trajectories are associated by a hand-designed association strategy, the errors of association will introduce error supervision into the system. Therefore, we propose to parallel a Kalman filter during the training process to generate pseudo-labels for those unlabeled data for semi-supervised training. The experimental results prove the effectiveness of our strategy.

Specifically, our contributions are as follows:
\begin{itemize}
\item We propose a data-driven MOT method by using a GRU-based motion module to avoid the precision loss of traditional methods for noise mismatch modeling and motion process linearization.
\item We design a pseudo-label-based semi-supervised method, which greatly expands the amount of available training data and label robustness, so that the system can converge in fewer training cycles.
\item We evaluate our method on the nuScenes\cite{caesar2020nuscenes} and Argoverse2\cite{wilson2023argoverse} datasets. Our system demonstrates performance comparable to traditional MOT systems and strong level of generalization, while obviating the need for manually designing a model for each object category.
\end{itemize}

\section{RELATED WORK}

Lidar-based 3D MOT has a similar form to image-based MOT. However, the lidar point cloud provides more precise spatial and depth information, significantly enhancing tracking precision. This improvement has led to its widespread adoption in robotic sensing and other applications. Most Lidar-based 3D MOT algorithms can be categorized into two primary paradigms: ``Tracking-by-Detection'' and ``Joint Detection and Tracking''. 

\subsection{TBD}

Leveraging increasingly advanced detectors, AB3DMOT\cite{weng20203d} establishes a baseline method for 3D MOT based on filters and 3D Intersection over Union (3D IoU), which provides a foundation for the design of TBD methods. 

%The TBD framework comprises four key steps: (1) preprocessing 3D detection results; (2) predicting and updating the motion of existing object trajectories; (3) matching predicted trajectories with detected objects; and (4) managing the life cycle of trajectories.

SimpleTrack\cite{pang2022simpletrack} offers a detailed analysis of the strengths and limitations of various models, including the widely used Kalman filter and the constant velocity model, as well as different data association metrics. It proposes specific improvements for each module and integrates them into a streamlined baseline method, achieving competitive results on the Waymo Open Dataset and nuScenes.

Additionally, Poly-MOT\cite{li2023poly} incorporates geometric constraints into the motion model and develops multiple motion models tailored to the distinct characteristics of different object categories. It introduces three custom similarity measures and a novel two-stage data association strategy, enabling the identification of the most suitable similarity measure for each object category and reducing mismatches. This approach further optimizes TBD trackers, resulting in superior tracking performance on the NuScenes dataset.

Some methods aim to improve tracking performance by integrating image features. For instance, EagerMOT\cite{kim2021eagermot} combines image detection results with Lidar detection data to achieve comprehensive scene perception. CAMO-MOT\cite{wang2023camo} also leverages both camera and Lidar data to significantly mitigate tracking failures caused by occlusions and false detections. Additionally, CAMO-MOT introduces an occlusion head to effectively select optimal object appearance features multiple times, further reducing the impact of occlusions.

\subsection{JDT}

Motiontrack\cite{zhang2023motiontrack} proposed an end-to-end Transformer-based\cite{vaswani2017attention} JDT algorithm, which was based on the previous end-to-end detection work Transfusion\cite{bai2022transfusion}, and further proposed a Transformer-based data association module and query enhancement module. In addition, following Guillem et al.\cite{braso2020learning}, OGR3MOT\cite{zaech2022learnable} uses graph neural networks to solve the 3D MOT problem, and achieves the best IDS metrics so far.

\subsection{Recursive Bayesian Filtering}

In MOT, predicting and updating existing trajectories is a crucial aspect. KF\cite{Kalman} is used for linear Gaussian state models. However, in practice, many problems do not perfectly adhere to the linear Gaussian model. Consequently, nonlinear filters, such as the EKF \cite{Gruber}, are employed for approximate processing. It performs recursive computation by linearizing the state forward function and the observation function.

The model-based algorithms aforementioned, which rely on accurate knowledge of the SS model, often experience significant performance degradation when there is a mismatch between the actual motion model and the modeled one\cite{redmon2016you}. Recently, there has been an increasing focus on integrating machine learning with SS models. DNN-based algorithms typically encode observations to fit a simplified SS model, and then track the parameters of these implicit SS, as exemplified by KFNet \cite{zhou2020kfnet}. Additionally, some approaches\cite{Satorras_Akata_Welling_2019} incorporate graph neural network (GNN) alongside Kalman filters to enhance filter accuracy through neural augmentation. However, such algorithms are often designed for unknown or highly complex SS models that lack mathematical interpretability and are generally not suited for real-time estimation due to their computational demands.

In contrast, KalmanNet \cite{revach2022kalmannet} offers a novel approach by combining model-based Kalman filtering with RNN to address model mismatch and nonlinearity. Our work builds upon KalmanNet, applying it to practical MOT scenarios. This approach eliminates the need for manual design of multiple SS models and the selection of various filters for different classes, while avoiding additional computational complexity.

\section{METHOD}
\subsection{3D MOT Pipeline}
Our system can be divided into four parts: the pre-processing module, motion module, association module, and trajectory management module, as shown in Fig. \ref{fig:Overview}.

\subsubsection{Pre-processing Module}

3D MOD typically generates multiple bounding boxes for the same detection to minimize the risk of missed and false detections. To prevent these detections from causing redundant ID switches, we preprocess the original detections $D_n^{\prime}$ to reduce false matches. This preprocessing generally involves applying score filtering and non-maximum suppression (NMS) to each frame’s detections, retaining only the bounding box with the highest confidence for each object. After preprocessing, our detection boxes $D_n = [x,y,z,w,l,h,v,\theta]$ include the center position of the bounding box, its dimensions, its velocity, and the heading angle.

\subsubsection{Motion Module} The motion module is primarily responsible for predicting the state $\hat{X}_{n-1} = \left[ \hat{x}_n^1, \ldots, \hat{x}_n^{num\_tra} \right]$ of the tracked trajectories and updating the state based on observations $Y_n= \left[y_n^1, \ldots, y_n^{num\_det} \right]$. Our design focuses on this module.

The process of traditional motion and prediction, taking EKF as an example, is shown in Fig. \ref{fig:EKF}. The system continuously updates the state through the prior state $\hat{x}_{n-1}$, the observation $y_n$, the process noise covariance matrix $Q$, the measurement noise covariance matrix $R$, and the continuously maintained state covariance $P$. Specifically, there are two steps:

\begin{figure}[tbp]
    \centering
	\includegraphics[width=1\linewidth]{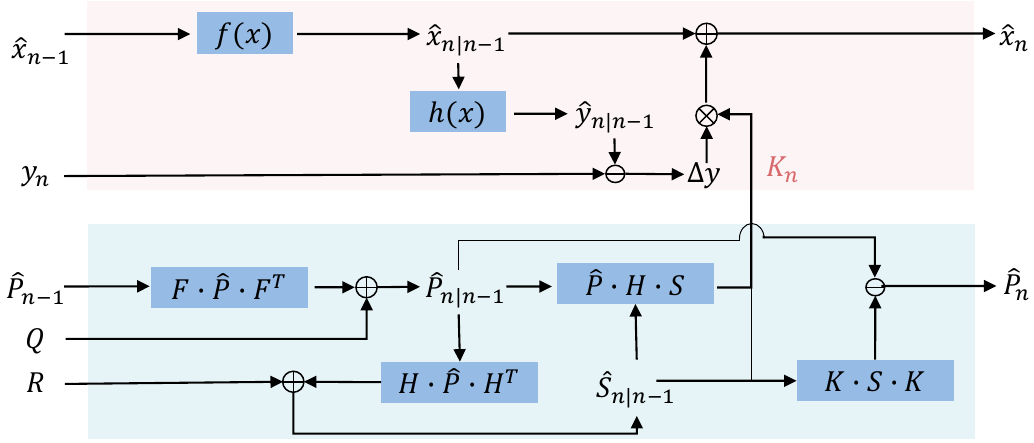}
    \caption{\textbf{EKF Block Diagram.}}
    \label{fig:EKF}%文中引用该图片代号
    \vspace{-5mm}
\end{figure}

a) Prediction Step: In this step, the system predicts the current prior state based on the posterior state from the previous time step, following the predefined SS model:
\begin{equation}
\hat{x}_{n \mid n-1}=f(\hat{x}_{n-1}).
\label{eq.1}
\end{equation}
Simultaneously, the uncertainty in the state, represented by the covariance, must also be predicted:
\begin{equation}
\hat{P}_{n \mid n-1}=F \cdot \hat{P}_{n-1} \cdot F^T + Q.
\label{eq.2}
\end{equation}

b) Update Step: The first step involves calculating the Kalman gain $K_n$, which balances the weights between the prediction and the observation:
\begin{equation}
K_n=\hat{P}_{n \mid n-1} \cdot H^T \cdot(H \cdot \hat{P}_{n \mid n-1} \cdot H^T + R)^{-1},
\label{eq.3}
\end{equation}
\begin{equation}
\hat{S}_{n \mid n-1}=H \cdot \hat{P}_{n \mid n-1} \cdot H^T + R ,
\label{eq.4}
\end{equation}
where $\hat{S}_{n \mid n-1}$ is the observation error covariance.

Again according to the current observation, update the current state of posterior:
\begin{equation}
\hat{x}_n=\hat{x}_{n \mid n-1} + K_n \cdot (y_n-h(\hat{x}_{n \mid n-1})).
\label{eq.5}
\end{equation}
Meanwhile, the error covariance is updated as follows: 
\begin{equation}
\hat{P}_n= \hat{P}_{n \mid n-1} - K_n \cdot \hat{S}_{n \mid n-1} \cdot K^T_n.
\label{eq.6}
\end{equation}
Here, the EKF  uses the Jacobian matrix $F$ and $H$ to linearize the differentiable functions $f(x)$ and $h(x)$ in a time-dependent manner.

This approach relies heavily on the accuracy of the SS model setup, which often depends on the designer's experience and is difficult to transfer. Moreover, for nonlinear motion, the linearization of the state transition and observation equations can introduce additional errors. Additionally, in MOT, the Kalman filter requires manual adjustment of the observation noise and process noise, assuming they follow a multi-dimensional Gaussian distribution. However, in MOT, observations are derived from upstream object detection results, which do not necessarily conform to a Gaussian distribution. Thus, explicitly modeled noise may not adequately address the needs of MOT.

It is reasonable to believe that the differences in motion for different classes are often also reflected in the state. If the model can adaptively obtain different Kalman gain results based on the observations and state, it can then be applied to all classes and adjust the noise accordingly.

\begin{figure}[tbp]
    \centering
	\includegraphics[width=1\linewidth]{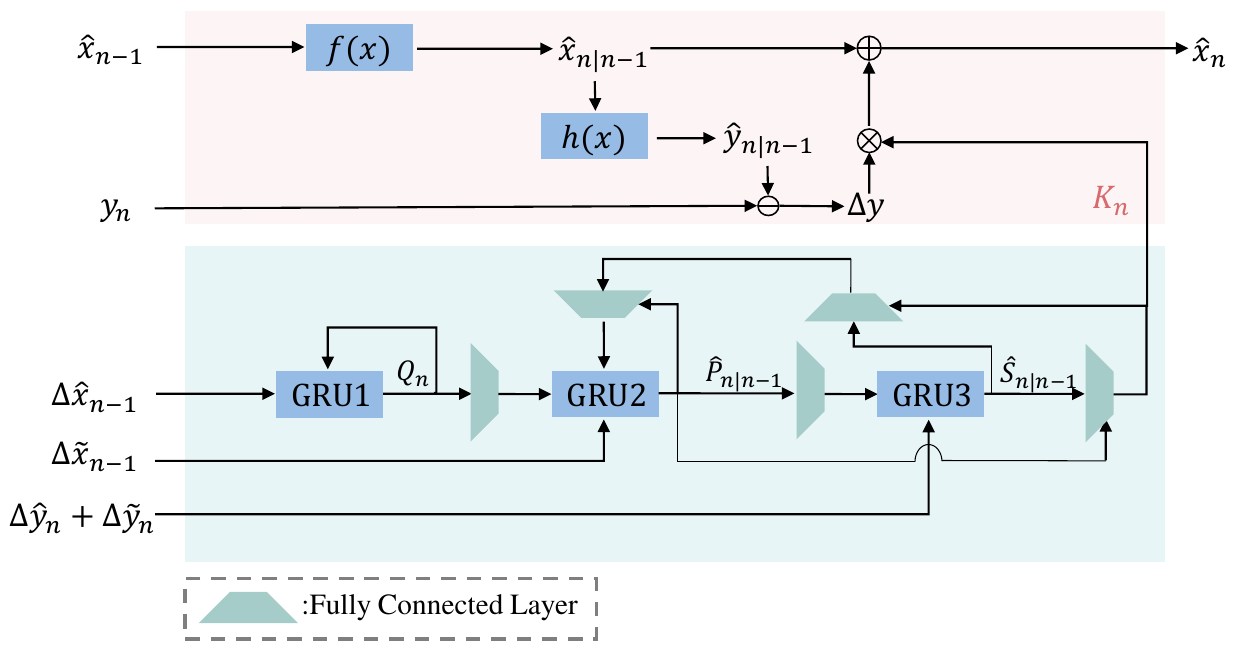}
    \caption{\textbf{The GRU-Kalman Filter Block Diagram.} Here, GRU simulates the loop iteration of process noise $Q_n$, state error covariance $\hat{P}_{n \mid n-1}$ and observation error covariance $\hat{S}_{n \mid n-1}$ by inputting the state difference and observation difference, so as to reason about the Kalman gain $K_n$.}
    \label{fig:GRU-KalmanNet}%文中引用该图片代号
    \vspace{-5mm}
\end{figure}

\textbf{GRU-Kalman Filter:} We introduce the learnable Kalman filter\cite{revach2022kalmannet} into our system, using a uniform SS model across all classes. The specific neural network architecture is illustrated in Fig. \ref{fig:GRU-KalmanNet}. Following the design of KalmanNet, we represent each second-order statistical moment of the Kalman filter using separate GRU, with fully connected (FC) layers interspersed between the GRU, and dedicated input and output layers.

%In addition, for MOT, the use of Kalman filter needs to manually adjust the Process noise $Q$ and the Measurement noise $R$, and it is assumed that it conforms to the multi-dimensional Gaussian distribution. In the subsequent manual adjustment process, $Q$ and $R$ are manually adjusted to make the index of the whole system achieve the best. However, in multi-object tasks, the observations come from the results of upstream object detection, which cannot be well assumed to conform to a Gaussian distribution. So, explicit noise does not meet the needs of multi-objective tasks.

%For different classes, the difference of their motion is often reflected in the observation. If the model can adaptively obtain different Kalman gain results with the observation data, one model can be used to adapt to all classes.

%\textbf{GRU-KalmanNet:} We introduce the learnable Kalman filter KalmanNet into the system, and uniformly use one SS model for each class. The specific neural network architecture is shown in FIG. \ref{fig:GRU-KalmanNet}. Following the KalmanNet, we represent each second-order statistical moment of the Kalman filter by a separate GRU unit, and some FC layers are interspersed between each GRU, and there are dedicated input and output layers.

Specifically, we use the first GRU to track the process noise covariance matrix, which combines the previous process noise covariance matrix $Q_{n-1}$ and the state forward update difference $\Delta \hat{x}_n=\hat{x}_n - \hat{x}_{n \mid n-1}$ to infer the current process noise covariance matrix $Q_n$.  

The second GRU is employed to simulate the reasoning process of Eq. \ref{eq.2}, thereby circumventing the need to linearize $f(x)$ in order to design $F$. It combines the process noise covariance matrix and the forward evolution difference $ \Delta \Tilde{x}_n=\hat{x}_n - \hat{x}_{n-1}$, along with the previous state error covariance $\hat{P}_{n-1 \mid n-2}$, to infer the current state error covariance $\hat{P}_{n \mid n-1}$. 

 The third GRU is designed to simulate the reasoning process of the observation formula in Eq. \ref{eq.4}, avoiding the linearization of $h(x)$ required to design $H$. It integrates the state covariance $\hat{P}_{n \mid n-1}$, the observation difference$\Delta \Tilde{y}_n= y_n - y_{n-1}$, the innovation difference $\Delta y_n = y_n - \hat {y}_{n \mid n-1}$, and the previous observation error covariance $\hat{S}_{n-1 \mid n-2}$ to derive the current observation error covariance $\hat{S}_{n \mid n-1}$. 
 
Finally, simulating the traditional Kalman filter process, we learn the $H$ matrix through an output layer, which takes the state error covariance and observation error covariance as inputs and ultimately outputs the Kalman gain $K_n$. Compared with end-to-end DNN, this network structure emulates part of the model-based Kalman filter process, making it more parameter-efficient and easier to train.

\begin{table*}[htbp]
    \centering
    \renewcommand{\arraystretch}{1.2} % 设置每行的间距为 1.5 倍
    \caption{A Comparison Of Existing Methods Applied To The Nuscenes Test Set.}
    \begin{threeparttable}
    \setlength{\tabcolsep}{6mm}{
    \begin{tabular}{cccccc}
    \toprule[1pt]
    Method & \textbf{Detector} & \textbf{Input Data} & \textbf{AMOTA$\uparrow$} & \textbf{AMOTP$\downarrow$} & \textbf{IDS$\downarrow$} \\ 
    \midrule
    CBMOT\cite{benbarka2021score} & CenterPoint\cite{yin2021center} \& CenterTrack\cite{zhou2020tracking} & 2D + 3D & 68.1 & 52.8 & 761  \\ 
    EagerMOT\cite{kim2021eagermot} & CenterPoint \& Cascade R-CNN\cite{cai2018cascade} & 2D + 3D & 67.7 & 55.0 & 1156  \\ 
    SimpleTrack\cite{pang2022simpletrack} & CenterPoint  & 3D & 66.8 & 55.0 & 575  \\ 
    OGR3MOT\cite{zaech2022learnable} & CenterPoint  & 3D & 65.6 & 62.0 & 288  \\ 
    CenterPoint & CenterPoint  & 3D & 65.0 & 53.5 & 684  \\ 
    Poly-MOT\cite{li2023poly} & CenterPoint  & 3D & 70.0 & 50.9 & 331  \\ 
    {Poly-MOT\tnote{1}} & LargeKernel3D\cite{chen2022scaling}  & 2D + 3D & 75.4(74.2) & 42.2(46.5) & 292(\textbf{276})  \\ 
    \midrule
    Ours & CenterPoint  & 3D & 70.0 & 50.4 & 353  \\ 
    Ours & LargeKernel3D  & 2D + 3D & \textbf{74.4} & \textbf{42.9} & 352  \\ 
    \bottomrule[1pt]
    \end{tabular}
    }
    \end{threeparttable}
    \label{tab:comparison_test}
\begin{tablenotes}    %这行要添加， 从这开始
    \footnotesize               %这行要添加
    \item{1} ``()'' indicates the result we reproduced.         %这行要添加
\end{tablenotes}            %这行要添加
\vspace{-5mm}
\end{table*}

\subsubsection{Association Module}

%Following the baseline Poly-MOT, we implement the data repeat association module, which reduces false negative matches through a two-stage association strategy that employs different similarity measures.

%\textbf{One-stage Association:} After obtaining through motion prediction from the existing trajectory $T_{n-1}$, we need to perform one-to-one matching between $T_{n-1}$ and detections $D_n$. Specifically, we follow the baseline and use the 3D Generalized Intersection over Union ($GIoU_{3D}$) \cite{rezatofighi2019generalized} to compute the affinity between them.

%where $Hull$ denotes the convex hull in the space of both objects. Compared to IoU and Euclidean distance, it combines the advantages of both, allowing it to handle sudden position changes while also distinguishing object orientations. Finally, we use the Hungarian algorithm to match $T_{n-1}$ and detections $D_n$.

%\textbf{Two-stage Association:} After the one-stage matching, the mismatched trajectories$T^u_{n-1}$ and detections $D^u_n$ can lead to unnecessary trajectory termination and initialization within the system. To mitigate this false negative error, we employ a two-stage matching process. In this stage, we use GIoU in the Bird’s Eye View ($GIoU_{BEV}$) to compute the affinity, and again apply the Hungarian algorithm to match with detections. 

%By combining the two-stage matching, we effectively obtain three outcomes: updated trajectories $T_n^D$, mismatched trajectories$T^u_{n}$, and mismatched detections $D^u_n$.

Following the baseline Poly-MOT, we use a two-stage association strategy to reduce false negative associations. In the One-stage Association, we use the 3D Generalized Intersection over Union ($GIoU_{3D}$) to as the match metric between the tracked trajectories and the detections. After One-stage Association, we perform a wider threshold Two-stage Association between mismatched trajectories and mismatched detections. This association will be performed in Bird’s Eye View.
\begin{equation}
\operatorname{GIoU_{3D}}\left(T_{n-1}, D_n\right) = \frac{T_{n-1} \cap D_n}{T_{n-1} \cup D_n}+\frac{T_{n-1} \cup D_n}{{H u l l}\left(T_{n-1}, D_n\right)}-1.
\label{eq.7}
\end{equation}

By combining the two-stage association, we effectively obtain three results: updated trajectories $T_n^D$, mismatched trajectories$T^u_{n}$, and mismatched detections $D^u_n$.

\subsubsection{Trajectory Management Module}
Like most 3D-MOT methods, our system employs a confidence-based trajectory lifecycle management approach\cite{benbarka2021score}. Specifically, when the tracker receives a detection $D^u_n$ that is not associated with an existing trajectory, it is initialized as a new trajectory. During subsequent tracking, if a observation is associated with this trajectory, corresponding to $T^D_n$, the state of the trajectory is updated based on the new observation. However, if the trajectory is not observed in consecutive frames$T^u_{n-1}$, its confidence decays following an exponential function until it is ultimately deleted.

\subsection{Training methods}
Our design also extends to the training methodology. The prototype system, KalmanNet, is trained in a supervised manner using labeled datasets. The squared error is computed between the estimated value $\hat{X}_n$ and the true value $X_n$, as shown below:
\begin{equation}
\mathcal{L}=\sum_{n=0}^{seq}\left\|{X}_n-\hat{{X}}_n\right\|^2.
\label{eq.8}
\end{equation}
Although the system output is the Kalman gain, supervising with respect to the state variable $X_n$ is not direct supervision. However, it is still effective because:
\begin{equation}
\begin{aligned}
\frac{\partial \mathcal{L}_n}{\partial {K}_n} & =\frac{\partial\left\|{K}_n \Delta \mathbf{y}_n-\Delta {X}_n\right\|^2}{\partial {K}_n} \\
& =2 \cdot\left({K}_n \cdot \Delta {y}_n-\Delta {X}_n\right) \cdot \Delta {y}_n^{\top}
\end{aligned},
\label{eq.9}
\end{equation}
where $\Delta {X}_n={X}_n-\hat{X}_{n \mid n-1}$. The loss is differentiable with respect to the Kalman gain, meaning that KalmanNet can be trained end-to-end to compute the Kalman gain by minimizing the squared error.

However, in KalmanNet, the training data is automatically generated and can be collected in unlimited quantities. Automatically generated labels ensure a strict correspondence between labels and samples. However, in MOT, the annotations is provided by the labeled bounding boxes in the dataset. On the one hand, the number of annotations labels is typically smaller than the number of objects being tracked. On the other hand, associating annotations bounding boxes with trajectories often relies on methods such as Euclidean distance or IoU. Regardless of the association method used, there is no guarantee that the annotations strictly correspond to the system's trajectories, which brings anomalous supervision to the system.

To address these issues, we adopt a semi-supervised training method based on pseudo-labels. Specifically, in addition to using the annotations to provide the true state values for supervised training, we also parallelize the training process with an EKF. For trajectories that are not associated with annotated boxes, we use the system state obtained by the EKF as a pseudo-label for training. Consequently, our final loss function is as follows:
\begin{equation}
\mathcal{L}=\sum\left\|{X}_n-\hat{{X}}_n\right\|^2+\sum\left\|\tilde{{X}}_n-\hat{{X}}_n\right\|^2,
\label{eq.10}
\end{equation}
where ${X}_n$ is the true annotation of the dataset, $\tilde{{X}}_n $is the update output of traditional Kalman filter, and $\hat{{X}}_n$ is the update output of GRU-Kalman filter.

\begin{table*}[t]
    \centering
    \renewcommand{\arraystretch}{1.2} % 设置每行的间距为 1.5 倍
    \caption{A Comparison Of On The Argoverse2 Val Set.(HOTA)}
    \label{tab:comparison Argoverse2 val set}
    \resizebox{\textwidth}{!}{
    \begin{tabular}{cccccccccc}
    \toprule[1pt]
        ~ & \textbf{Detector} & \textbf{REGULAR\_VEHICLE} & \textbf{PEDESTRIAN} & \textbf{BICYCLE} & \textbf{LARGE\_VEHICLE} & \textbf{BUS} & \textbf{BOX\_TRUCK} & \textbf{TRUCK} & \textbf{Average}  \\ 
    \midrule
        Greedy & LT3D\cite{peri2023towards} & 58.9 & 59.1 & 51.6 & 28.5 & 48.2 & 44.0 & 31.3 & 46.0  \\ 
        AB3DMOT & LT3D & 59.2 & 54.6 & 48.7 & 26.7 & 47.0 & 43.3 & 34.2 & 42.7  \\ 
        Ours & LT3D & \textbf{73.3} & \textbf{71.9} & \textbf{53.8} & \textbf{29.3} & \textbf{59.2} & \textbf{55.3} & \textbf{36.0} & \textbf{47.3}  \\ 
    \bottomrule[1pt]
    \end{tabular}
    }
    \vspace{-5mm}
\end{table*}

\section{EXPERIMENTS}
\subsection{The dataset}
\subsubsection{NuScenes}
The nuSceness\cite{caesar2020nuscenes} dataset comprises 850 training sequences and 150 validation sequences, each consisting of approximately 40 frames. Keyframes are sampled at a rate of 2Hz, with annotations provided for each keyframe. These annotations include geometric details of the object bounding boxes and their unique identifiers within the scene. The official evaluation primarily employs accuracy AMOTA as the key performance metric, precision AMOTP, ID Switch (IDS) as the secondary performance metric.
\subsubsection{Argoverse2}
Argoverse2\cite{wilson2023argoverse} expands on Argoverse1\cite{Argoverse} by collecting 1000 scene clips in six US cities. Each sequence lasted 15 seconds, sampled and annotated at 10Hz, with an average of 75 annotated objects per frame. The dataset features over 30 object classes and encompasses multiple complex urban environments. The official evaluation uses HOTA as the key metric.

\subsection{Implementation Details}
The system input consists of the detection results from the 3D MOD. For dataset splitting, we follow the official division of training, validation, and test sets. Regarding training parameters, we employ the AdamW optimizer with a maximum learning rate 1e-5 and weight decay 1e-5. The CosineAnnealingLR adjustment method and smooth-loss supervision are applied to help the model avoid local minima. Gradient accumulation is performed across each object pair per frame in the sequence, followed by backpropagation at the end of the sequence.
\subsection{Experimental Results}
\subsubsection{Comparative evaluation: }We compare our approach with several methods on the validation and test sets of the NuScenes dataset, as well as on the validation set of the Argoverse2 dataset.

\textbf{NuScenes Test Set:} We compared our system with other tracking algorithms, such as the TBD baselines AB3DMOT, CenterTrack, SimpleTrack, Poly-MOT, and JDT method OGR3MOT. The selected algorithms include both traditional tracking methods and machine learning-based approaches.

As shown in TABLE \ref{tab:comparison_test}, according to the AMOTA results, our proposed GRU-based semi-supervised method achieves better performance than most mainstream hand-designed methods using the same detector (CenterPoint\cite{yin2021center}). Compared to OGR3MOT with GNN, it also demonstrates superior performance. Furthermore, we obtain the best AMOTP results, indicating that our system provides more accurate final trajectory box information than the traditional Kalman model and constant velocity model. Overall, compared with most TBDs and JDTs, we exhibit excellent performance without the need for manually designing motion models, highlighting great potential for future development. 

\begin{table}[ht]
    \centering
    \renewcommand{\arraystretch}{1.2} % 设置每行的间距为 1.5 倍
    \caption{A Comparison Of Existing Methods Applied To The Nuscenes Val Set.}
    \begin{tabular}{ccccc}
    \toprule[1pt]
    \textbf{Method} & \textbf{Input Data} & \textbf{AMOTA$\uparrow$} & \textbf{AMOTP$\downarrow$} & \textbf{IDS$\downarrow$} \\ 
    \midrule
        CBMOT & 2D + 3D & 72.0 & 48.7 & 479  \\ 
        EagerMOT & 2D + 3D & 71.2 & 56.9 & 899  \\ 
        SimpleTrack & 3D & 69.6 & 54.7 & 405  \\ \
        OGR3MOT & 3D & 69.3 & 62.7 & \textbf{262}  \\ 
        CenterPoint & 3D & 66.5 & 56.7 & 562  \\ 
        Poly-MOT & 3D & 73.1 & 52.1 & 281  \\ \midrule
        Ours & 3D & \textbf{73.2} & \textbf{51.1} & 267  \\  %6000steps
    \bottomrule[1pt]
    \end{tabular}
    \label{tab:comparison_val}
\end{table}

\textbf{NuScenes Val Set:} As shown in Table \ref{tab:comparison_val}, we evaluated our system on the validation set using the same model configuration, and once again, it demonstrated stable and consistent performance.

\textbf{Argoverse2 Val Set:} We evaluated our method on the Argoverse2 validation set in Table \ref{tab:comparison Argoverse2 val set}, which includes ground-truth labels for 30 object classes. Given that most MOT algorithms require modification before being applied to Argoverse2, we compared our method with the Argoverse2 baseline, LT3D, using tracking method such as greedy and ab3dmot tracker. We present results for the seven most prevalent dynamic object categories in Argoverse2 and report the overall average HOTA score. Importantly, we used the model trained on the NuScenes training set, and its performance on Argoverse2 demonstrates a strong level of generalization.

%在这一部分中，我们进行了广泛的消融实验，以评估所提出的模块的性能。我们使用了相同的预处理参数和关联参数，然后更换motion module以及不同训练策略进行了一系列实验。

\subsubsection{Ablation Study}In this part, we conduct extensive ablation experiments to evaluate the performance of the proposed module. We used the same preprocessing parameters and association parameters, and then replaced the motion module as well as different training strategies for a series of experiments.

%1）我们选取了最具有代表性的两个类别，单独考察不同的SS和优化器在这两个类别上的跟踪性能表现。从表结果可以看出，以Poly-MOT为代表的传统方法，在不同SS模型和优化器下性能会有所出入，这正是传统的运动模块高度依赖准确的建模所致。而我们的基于数据驱动的方法，在两类中都取得了最好的跟踪结果。并且更改SS并不会为跟踪精度带来影响，这体现了我们运动模块优异的鲁棒性。

\begin{table}[tpb]
    \centering
    \renewcommand{\arraystretch}{1.2} % 设置每行的间距为 1.5 倍
    \caption{The effectiveness ablation experiments were conducted on the Nuscenes validation set. Here, the CTRA, BICYCLE, and CA state models follow the PolyMOT specification.}
    \begin{tabular}{cccc}
    \toprule[1pt]
    \textbf{Category} & \textbf{Motion Module} & \textbf{AMOTA$\uparrow$} & \textbf{AMOTP$\downarrow$} \\ 
        \midrule
        ~ & Bicycle + EKF & 54.5 & 50.7 \\ 
        ~ & CTRA + EKF & 55.0 & 46.2 \\ 
        Bicycle & CA + KF & 54.5 & 46.3 \\ \cline{2-4}
        ~ & CTRA + Ours & \textbf{55.6} & \textbf{45.9} \\ 
        ~ & Bicycle + Ours & \textbf{55.6} & \textbf{45.9} \\ 
        \midrule
        ~ & Bicycle + EKF & 86.3 & 34.2 \\ 
        ~ & CTRA + EKF & 86.3 & 34.0 \\ 
        Car & CA + KF & 86.0 & \textbf{33.6} \\ \cline{2-4}
        ~ & CTRA + Ours & \textbf{86.3} & 33.8 \\ 
        ~ & Bicycle + Ours & \textbf{86.3} & 33.8 \\ 
    \bottomrule[1pt]
    \end{tabular}
    \label{tab:comparison_model}
\end{table}

\textbf{GRU-based Motion Module:} We select the two most representative categories to investigate the tracking performance performance of different SS and optimizers on these two categories separately. As can be seen from the Table \ref{tab:comparison_model}, the performance of the traditional methods represented by Poly-MOT will vary under different SS models and optimizers, which is precisely caused by the high dependence of the traditional motion module on accurate modeling. However, our data-driven based method, achieves the best tracking results in both categories. And changing SS does not affect the tracking accuracy, which reflects the excellent robustness of our motion module.

   \begin{figure}[tpb]
      \centering
        \includegraphics[width=1\linewidth]{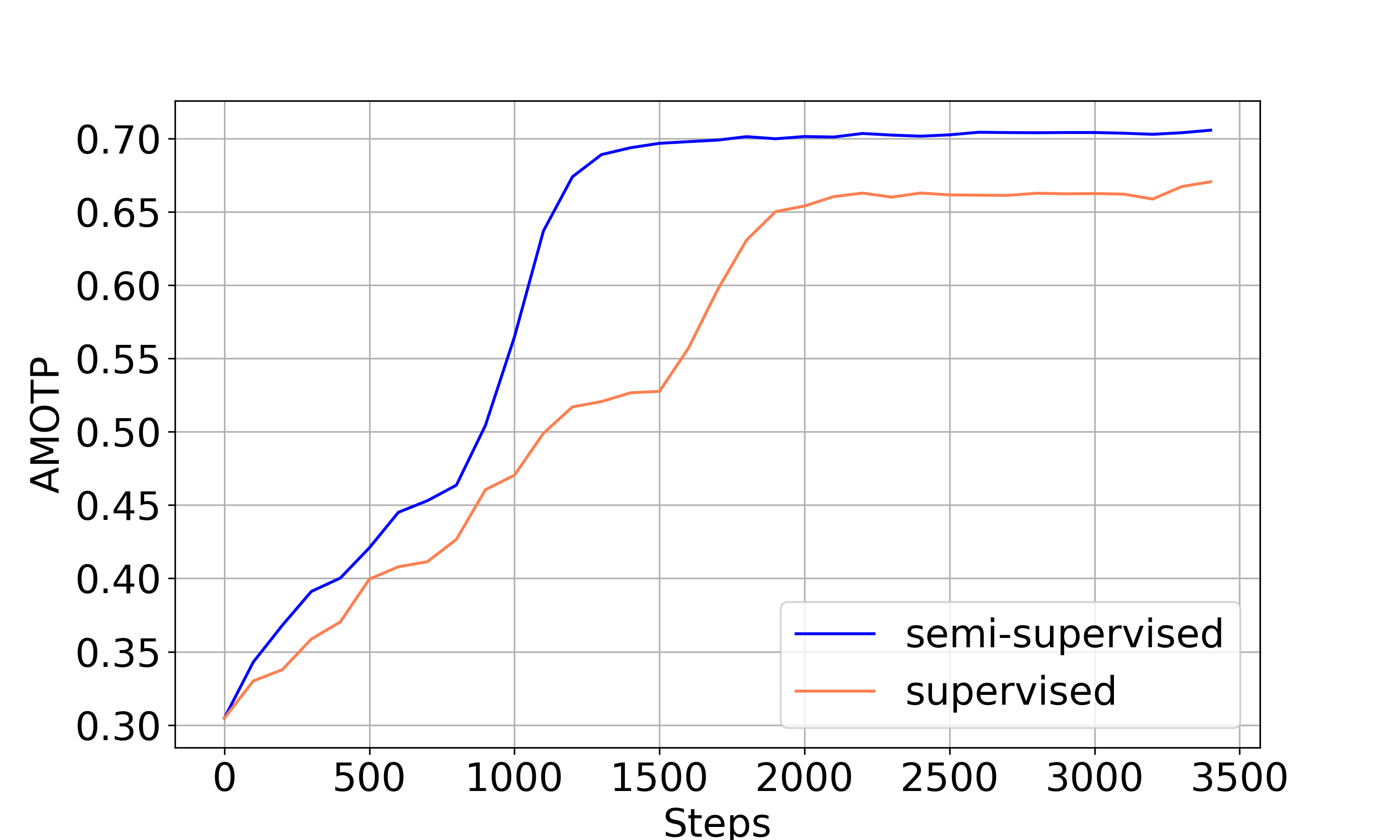}
      \caption{Experiments comparing the training convergence speed and accuracy of supervised and semi-supervised training.}
      \label{Ablation Study}
   \end{figure}

\textbf{Semi-Supervised:} To verify the effectiveness of our proposed semi-supervised method, we compared the fully supervised and semi-supervised training processes and results on the validation set. As shown in Figure \ref{Ablation Study}, our semi-supervised approach achieved convergence within 1,700 steps, or roughly 2 epochs, reaching an AMOTA score close to 0.7. In contrast, when training solely with only annotations from dataset, the system took 3 epochs to converge, and the final performance was lower. This discrepancy is attributed to the limited amount of training data and errors introduced during the association between annotations and the tracked samples, which semi-supervised learning helps to mitigate.

\section{CONCLUSIONS}
In this work, we propose a partially learnable MOT method by introducing a GRU-based Kalman filter\cite{revach2022kalmannet} into the motion module of the TBD framework. This method eliminates the adverse effects of inaccurate noise parameterization and reduces the error of linearization. Our findings demonstrate that the GRU-based motion module is well-suited for MOT in autonomous driving and robotic sensing environments. Additionally, we introduce a semi-supervised training strategy that leverages pseudo-labels, which accelerates training by increasing data volume and minimizing association errors. We believe that combining deep learning methods with interpretable mathematical models can enhance 3D MOT performance, and foresee extending this approach to other modules, potentially leading to the development of parameter-free trackers.
%\begin{thebibliography}{99}
\bibliographystyle{IEEEtran}
\bibliography{IEEEabrv,mylib}

%\end{thebibliography}

\end{document}